\documentclass[acmsmall]{acmart}
\usepackage{makecell}
\usepackage{multirow}
\usepackage{tabularx,booktabs}
\usepackage{url}
\usepackage{bm}
\usepackage{hyperref}
\hypersetup{colorlinks}

\usepackage{subcaption}

\usepackage{cleveref}
\usepackage{algorithm}
\usepackage{algpseudocode}
\algdef{SE}[SUBALG]{Indent}{EndIndent}{}{\algorithmicend\ }%
\algtext*{Indent}
\algtext*{EndIndent}

\usepackage{graphicx}
\usepackage{amsmath}
\usepackage{booktabs}
\usepackage{enumitem}
\usepackage{comment}
\usepackage{bm}
\usepackage{multirow}

\newtheorem{prop}{Proposition}
\AtBeginDocument{%
  }

\setcopyright{none}

\acmConference[Preprint]{}{April}{2024}

\usepackage{cleveref}

\begin{document}

\title{AdaContour: Adaptive Contour Descriptor with Hierarchical Representation}


\author{Tianyu Ding}
\affiliation{%
  \institution{Microsoft Corporation}
  \streetaddress{One Microsoft Way}
  \city{Redmond}
  \state{Washington}
  \country{USA (tianyuding@microsoft.com)}
  \postcode{98052}}
\email{tianyuding@microsoft.com}

\author{Jinxin Zhou}
\affiliation{%
  \institution{Ohio State University}
  \streetaddress{281 W Lane Ave}
  \city{Columbus}
  \state{Ohio}
  \country{USA (zhou.3820@buckeyemail.osu.edu)}
  \postcode{43210}}
  \email{Zhou.3820@buckeyemail.osu.edu}

\author{Tianyi Chen}
\affiliation{%
  \institution{Microsoft Corporation}
  \streetaddress{One Microsoft Way}
  \city{Redmond}
  \state{Washington}
  \country{USA (tiachen@microsoft.com)}
  \postcode{98052}}
\email{tiachen@microsoft.com}

\author{Zhihui Zhu}
\affiliation{%
  \institution{Ohio State University}
  \streetaddress{281 W Lane Ave}
  \city{Columbus}
  \state{Ohio}
  \country{USA (zhu.3440@osu.edu)}
  \postcode{43210}}
\email{zhu.3440@osu.edu}

\author{Ilya Zharkov}
\affiliation{%
  \institution{Microsoft Corporation}
  \streetaddress{One Microsoft Way}
  \city{Redmond}
  \state{Washington}
  \country{USA (zharkov@microsoft.com)}
  \postcode{98052}}
\email{zharkov@microsoft.com}

\author{Luming Liang}
\affiliation{%
  \institution{Microsoft Corporation}
  \streetaddress{One Microsoft Way}
  \city{Redmond}
  \state{Washington}
  \country{USA (lulian@microsoft.com)}
  \postcode{98052}}
\email{lulian@microsoft.com}

\renewcommand{\shortauthors}{Ding et al.}

\begin{abstract}
Existing angle-based contour descriptors suffer from lossy representation for non-starconvex shapes. By and large, this is the result of the shape being registered with a single global inner center and a set of radii corresponding to a polar coordinate parameterization. In this paper, we propose AdaContour, an adaptive contour descriptor that uses \emph{multiple local} representations to desirably characterize complex shapes. After hierarchically encoding object shapes in a training set and constructing a contour matrix of all subdivided regions, we compute a robust low-rank robust subspace and approximate each local contour by linearly combining the shared basis vectors to represent an object. Experiments show that AdaContour is able to represent shapes more accurately and robustly than other descriptors while retaining effectiveness. We validate AdaContour by integrating it into off-the-shelf detectors to enable instance segmentation which demonstrates faithful performance. The code is available at \url{https://github.com/tding1/AdaContour}.
\end{abstract}




\maketitle

\section{Introduction}\label{sec:intro}

\begin{figure}
    \centering
\includegraphics[width=0.975\textwidth]{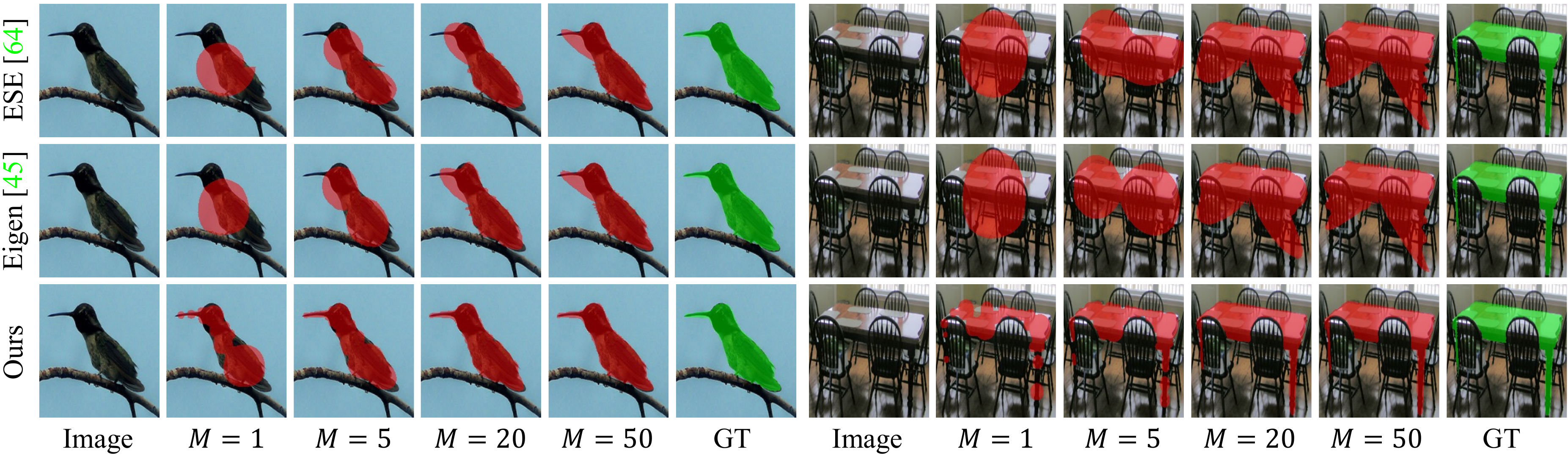}
\caption{\textbf{Two challenging examples in the SBD dataset~\cite{hariharan2011semantic}.} Existing dominant contour descriptors that use a \emph{single global} representation fail to capture subtle regions of the shapes, such as the beak of the bird and the legs of the table, for each choice of $M$ that controls the expressive ability. In sharp contrast, our approach with \emph{multiple local} representations yields high quality results.}\label{fig:teaser}
\end{figure}

Contour descriptors have a long history of use in graphics and vision applications, such as shape retrieval~\cite{pedronette2010shape,zhang2001comparative,alajlan2007shape} and image segmentation~\cite{xie2020polarmask,xu2019explicit,park2022eigencontours,peng2020deep,liu2021dance}, to represent shapes by outlining their boundaries. One of the greatest advantages of a contour descriptor is its compactness. The angle-based contour descriptors ~\cite{xu2019explicit,xie2020polarmask,park2022eigencontours}, for instance, register a shape with a single inner center, use polar coordinate parameterization with uniformly sampled angular coordinates, and identify the boundary points by the associated radial coordinates. Since the angle is inherently directional, a 1D sequence of radii is sufficient to determine the entire contour. However, these methods typically require the underlying shape to be (star-)convex; otherwise significant reconstruction errors would result (see~\Cref{fig:teaser}). 

Vertex-based contour descriptors~\cite{liu2021dance, peng2020deep,liang2020polytransform,yang2020dense}, on the other hand, utilize the Cartesian coordinate system and represent a shape as a collection of polygon vertex coordinates along the contour. Due to the 2D sampling sequences in the $x$- and $y$-axes, it is known that this representation is more sensitive to noise~\cite{xu2019explicit}, despite the fact that it can nearly fit any shape, including non-convex ones. When incorporated into learning-based algorithms for instance segmentation, it typically regresses per-vertex offsets to refine the contour towards the ground-truth object boundaries~\cite{peng2020deep,kass1988snakes,ling2019fast,liu2021dance}. Nevertheless, the uniformly sampled per-vetex regression leads to the problem of correspondence interlacing~\cite{liu2021dance}, which makes learning difficult and reduces its effectiveness.

\begin{figure}[t]
  \centering
  \begin{subfigure}{0.19\linewidth}
    \includegraphics[width=\textwidth]{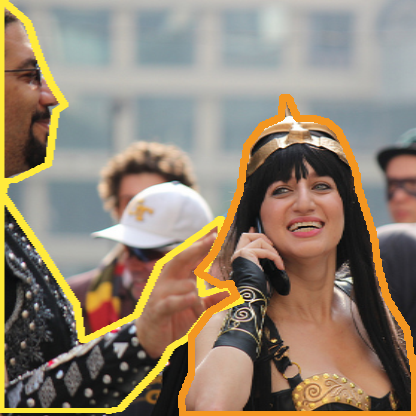}
    \caption{Ground-truth}
    \label{fig:short-a}
  \end{subfigure}
  \begin{subfigure}{0.19\linewidth}
    \includegraphics[width=\textwidth]{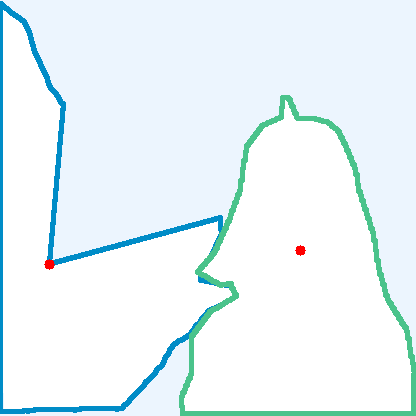}
    \caption{Eigen~\cite{park2022eigencontours}}
    \label{fig:short-b}
  \end{subfigure}
  \begin{subfigure}{0.19\linewidth}
   \includegraphics[width=\textwidth]{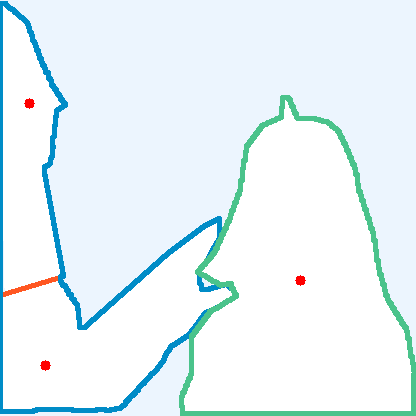}
    \caption{Ours (depth = 1)}
    \label{fig:short-c}
  \end{subfigure}
  \begin{subfigure}{0.19\linewidth}
    \includegraphics[width=\textwidth]{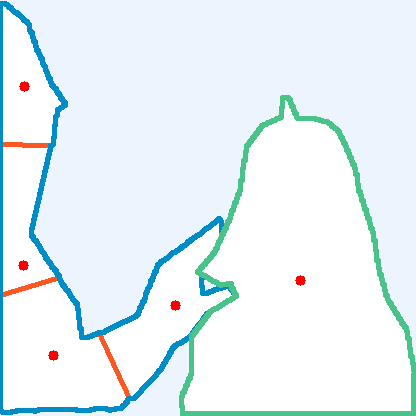}
    \caption{Ours (depth = 2)}
    \label{fig:short-d}
  \end{subfigure}
  \begin{subfigure}{0.19\linewidth}
    \includegraphics[width=\textwidth]{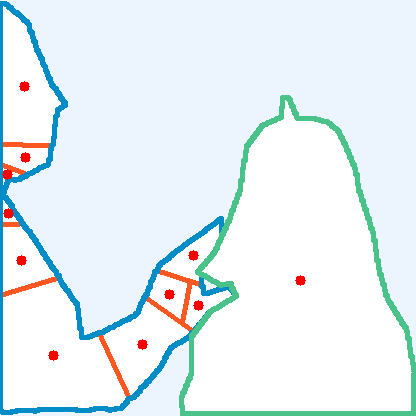}
    \caption{Ours (depth = 5)}
    \label{fig:short-e}
  \end{subfigure}
  \caption{\textbf{Illustration of hierarchical encoding.} (a) Image with ground-truth object boundaries. (b) Starconvex contours generated by~\cite{park2022eigencontours}. (c)-(e): Adaptive contours generated by our hierarchical encoding procedure, which recursively subdivides the initial possibly non-convex shape and terminates when a sufficiently regular area is identified or maximum depth is reached, then each subdivision is encoded by a local contour. 
  In the example, the object boundary of the lady is starconvex, rendering a single global contour sufficient for its representation. The man's boundary is more complex, and a single contour cannot faithfully represent the face, hand and chin. Hierarchical encoding is effective in resolving the issue by subdividing the shape adaptively and conquering each of them locally. By setting the depth to 1 and utilizing only 2 centers, the representation of the face and hand of the man is more precise. The hand shape becomes more distinctive at depth 2, and the depiction of the chin is rendered correctly at depth 5. Note that as the hierarchical depth increases, newly introduced local contours automatically converge around the challenging areas.} 
  \label{fig:dog}
\end{figure}

The focus of the present paper is \emph{AdaContour}, an adaptive angle-based contour descriptor with hierarchical representation that effectively addresses the aforementioned challenges. In contrast to the existing angle-based contour descriptors, which use a single global contour and cannot handle non-starconvex shapes, we use multiple local contours to characterize complex shapes in an adaptive manner. The key to success is a novel hierarchical encoding procedure (see~\Cref{fig:dog}) that recursively subdivides the shape until resulting in a sufficiently regular area or maximum depth is achieved, and then computes a local contour for each refined region. As a result, the identified local contours concentrate more on the region whose boundary exhibits rapidly varying curvatures, showing that hierarchical subdivision is adequately increasing the examination frequency along the irregular boundary. Here, angular coordinates are sampled uniformly inside each local contour.

Given a training set, we first \emph{hierarchically encode} all the object boundaries and construct a contour matrix by stacking together all the 1D vectors of radii of the local contours. We then compute a low-rank \emph{robust} subspace $\mathcal{S}$ for approximating the contour matrix. Instead of simply using SVD as in~\cite{park2022eigencontours}, which is known to be sensitive to outliers, we employ the more advanced robust subspace recovery method~\cite{lerman2018fast} to alleviate the issue. Finally, to efficiently represent an original object, each local contour is recovered by linearly combining the $M$ most dominating basis vectors of $\mathcal{S}$ (see Figure \textcolor{red}{1}). Note that the same set of $M$ basis vectors is \emph{shared} for reconstructing multiple local contours. Experiments show that AdaContour is superior to previous angle-based descriptors~\cite{park2022eigencontours,xie2020polarmask,xu2019explicit} due to its capability to neatly fit non-convex shapes. In addition, we incorporate AdaContour into object detectors YOLOv3~\cite{redmon2018yolov3} to enable instance segmentation and demonstrate faithful performance.

In summary, we make the following contributions:
\begin{itemize}
    \item We propose AdaContour, an adaptive contour descriptor that employs multiple local representations to effectively capture irregular shapes. This is the first time, to the best of our knowledge, that angle-based contour descriptors can precisely fit non-starconvex shapes.  

    \item We develop a novel hierarchical encoding algorithm that recursively determines the subdivisions of a given shape and encodes multiple local contours accordingly, which naturally concentrate around irregular areas that exhibit varying curvatures (see~\Cref{fig:dog}).  

    \item We utilize the robust subspace recovery approach to compute a low-rank robust subspace, which enables improved approximation of object boundaries with only a few coefficients. We also use basis-sharing conversion to enhance the efficiency of the recovery.

    \item We validate AdaContour by incorporating it into the YOLOv3 detector for instance segmentation. Our method, utilizing two local contours, exhibits favorable performance compared to conventional angle-based methods that utilize a single contour.
\end{itemize}

\section{Related work}
\label{sec:related-work}

The human visual system is highly effective at retrieving shape information from a dynamic and noisy environment. Every 3D object in the real world is perceived as a 2D shape in the retina, and is subsequently constrained by a 1D closed contour~\cite{koenderink1984does,elder2018shape}. Serving as such powerful visual cues, contour has numerous applications in modern vision problems, such as shape analysis~\cite{van1991contour, bai2009integrating,desimone1985contour,li2021multi,shu2011novel}, object detection~\cite{shotton2005contour,gong2018overview,arbelaez2010contour}, instance segmentation~\cite{xu2019explicit,xie2020polarmask,peng2020deep,liu2021dance,zhou2019cia,chen2017dcan}, and so on.

Given its importance, a great deal of effort has spent towards developing effective contour descriptors. \cite{van1991contour} proposes seven functions, such as tangent and curvature function, to parameterize contour representations as well as their Fourier series expansions. \cite{cinque1998shape} approximates segments of the contour by using the Bezier cubic curves. \cite{perez1994optimum} suggests fitting the ground-truth boundary points with a polygonal curve by optimizing the approximation error globally. In addition to these structural innovations, there are many other contour descriptors based on multi-scale curvature information~\cite{mokhtarian1992theory,dudek1997shape} and spectral transformation~\cite{zhang2001comparative,chuang1996wavelet}. For example,~\cite{chuang1996wavelet} uses wavelet transform to develop a hierarchical curve descriptor that decomposes a curve into multi-scale components, with the  coarsest scale components carrying the global approximation information and the finer components carrying local detailed information. 
Note that while it also uses hierarchical representation, it relies on a vertex-based wavelet descriptor that operates in the spectral domain, whereas the focus of this paper is the angle-based descriptor which is defined in the original planar domain.

Since convolutional neural networks have demonstrated success in many vision tasks, such as object detection~\cite{ren2015faster,girshick2015fast,gidaris2015object} and segmentation~\cite{he2017mask,minaee2021image,garcia2017review}, recent work resort to deep learning techniques to predict contours~\cite{bertasius2015deepedge,yang2016object,shen2015deepcontour}. Moreover, due to the strong connection between 1D contour and 2D shape, many contour-based instance segmentation methods have emerged~\cite{yang2020dense,liang2020polytransform,peng2020deep,liu2021dance}, most of which are originated from the conventional snake algorithms~\cite{kass1988snakes, gunn1997robust, cohen1991active, cootes1995active} that implicitly learn the contour boundary through vertex-based iterative refinements. Meanwhile, there are many other studies explicitly incorporate angle-based contour descriptors into object detection frameworks and enable one-stage instance segmentation by predicting the center and associated set of radii for generating the segmentation result. For this line of study,~\cite{xie2020polarmask} presents an anchor-free method by utilizing the object detector FCOS~\cite{tian2019fcos} to regress the radii at uniformly sampled angular directions in a centroidal profile. \cite{xu2019explicit} introduces an additional branch into the YOLOv3 detector~\cite{redmon2018yolov3} to regress the Chebyshev polynomial coefficients of the radii. Instead of representing a boundary by polynomial fitting,~\cite{park2022eigencontours} proposes a data-driven approach by analyzing boundary data in a training set and then efficiently represent object boundaries in a low-dimensional space by means of SVD. All of these studies use angle-based descriptors with a single global contour. Although they are computationally efficient, they suffer from lossy reconstructions for non-starconvex shapes.


This work also adopts a data-driven approach to leverage the distribution of object contours present in a training set. However, our proposed method surpasses existing techniques by employing hierarchical representation, robust subspace recovery techniques, and basis-sharing mechanisms, resulting in a more effective and efficient handling of non-starconvex shapes.

\section{The proposed approach}
\label{sec:approach}

\subsection{Background and formulation}\label{subsec:moti}

Existing angle-based contour descriptors register a shape using polar representation, where a single inner center is identified  by  either  using the mass center~\cite{xie2020polarmask} or selecting the point farthest from the contour via distance transform~\cite{maurer2003linear,xu2019explicit,park2022eigencontours}. Subsequently, a 1D sequence of radii (distance vector) corresponding to uniformly sampled angular coordinates is determined. For example,~\cite{xie2020polarmask} uniformly quantizes the 360-degree with an interval of $\Delta\theta=10^\circ$, resulting in a 36-dimensional distance vector for representing an object boundary. \cite{xu2019explicit,park2022eigencontours} utilizes $\Delta\theta=1^\circ$ to represent a boundary using a 360-dimensional vector. Despite their computational efficiency, they are unable to handle non-starconvex object boundaries adequately. This is due to their contour sampling mechanism that, if the ray casting from the inner center intersects the boundary more than once, only the point with the largest radius is adopted, resulting in inaccurate characterization of the shape.



Our approach to tackle this challenge begins with the observation that the object shape, despite its irregularity, can be subdivided into regular shapes, each of which can be adequately represented by a single local contour.  Given an input shape $\mathcal{M}$, our goal is to compute a subdivision $\Xi=\{s_1,\cdots,s_K\}$ such that each $s_i$ is sufficiently regular and can be efficiently represented by a simple local contour. To achieve this, we first quantify the regularity of a shape.  
We utilize  the concept of solidity~\cite{zdilla2016circularity} 
which is defined as the ratio of the area of an object to the area of its convex hull. The solidity of any shape is between 0 and 1;  the solidity of a convex shape is 1, and the larger the concavity, the smaller the solidity is.  
Therefore, we consider a region to be sufficiently regular if its solidity $\text{Sld}(\cdot)$ exceeds a certain threshold $\tau$. Based on this, our goal is to find a subdivision that maximizes the total solidities by solving
\begin{align}\label{eq}
    \max_{\Xi,K} \sum_{i=1}^K \text{Sld}(s_i)\quad \text{s.t.} \quad  K \le \overline K, \ \text{Sld}(s_i)\ge \tau, \ \forall i, \text{ and } 
    \bigcup \{s_i\} = \mathcal{M},
\end{align}
where $\overline{K}$ represents the maximum number of subdivisions.

\begin{algorithm}[t]
\caption{Hierarchical Encoding Procedure}
\label{alg:hie_encoding}
\begin{algorithmic}[1]
\small
\State \textbf{Input:} $\mathcal{M}\in \{0,1\}^{W\times H}$, $\tau\in(0,1)$, $D\in\mathbb{Z}^{\ge0}$, where $\mathcal{M}$ is a binary mask, $\tau$ is the solidity threshold, and $D$ is the maximum depth.  
\State \textbf{Output:} Set of centers $\mathcal{C}$ and set of distance vectors $\mathcal{R}$.
\State \textbf{Initialization:} $\mathcal{C}=\{\},\mathcal{R}=\{\}$.
\State \textbf{function} \textsc{Hierarchical-Encoding}($\mathcal{M},\tau,D$)
\Indent
\State  Calculate the solidity of the object in $\mathcal{M}$.
\If{$D=0$ or $\text{Sld}(\mathcal{M})>\tau$}\label{line:es}
\State Calculate the center $\bm c$ and distance vector $\bm r$.
\State Update $\mathcal{C}=\mathcal{C}\cup\{\bm c\}$ and $\mathcal{R}=\mathcal{R}\cup\{\bm r\}$.
\EndIf
\State Calculate the mass center $\bm c_{\mathcal{M}}$ of $\mathcal{M}$.
\State Find the direction $P$ of the least data variance in $\mathcal{M}$.\label{line:pca}
\State Calculate $\mathcal{M}_1,\mathcal{M}_2=$ \textsc{Split}($\mathcal{M},\bm c_{\mathcal{M}},P$).
\While{$\mathcal{M}_1$ or $\mathcal{M}_2$ is not connected}\label{line:s}
\State Update $\mathcal{M}_1,\mathcal{M}_2=$ \textsc{Reorg}($\mathcal{M}_1,\mathcal{M}_2$).\label{line:m}
\EndWhile\label{line:e}
\State \textsc{Hierarchical-Encoding}($\mathcal{M}_1,\tau,D-1$)
\State \textsc{Hierarchical-Encoding}($\mathcal{M}_2,\tau,D-1$)
\EndIndent
\State \textbf{end function}
\end{algorithmic}
\end{algorithm}

\subsection{Hierarchical encoding}\label{subsec:hie_enc}

As finding a direct solution to the constrained optimization problem~\eqref{eq} is challenging, we propose a hierarchical encoding procedure (\Cref{alg:hie_encoding}) in this section, which aims to solve it in a greedy manner. Given an object shape in the form of a binary mask\footnote{We assume input $\mathcal{M}$ is connected; otherwise (in a few number of circumstances), we deal with each of its connected components individually.} $\mathcal{M}\in\{0,1\}^{W\times H}$, our goal is to identify a set of local contours, or equivalently, a set of centers $\mathcal{C}$ and a set of radii $\mathcal{R}$, that can represent the object shape faithfully, where $\mathcal{C}=\{\bm c_i: \bm c_i\in\mathbb{R}^2\}_{i=1}^K,\mathcal{R}=\{\bm r_i : \bm r_i\in\mathbb{R}^N\}_{i=1}^K$, $K$ is the number of local contours, and $N$ is the dimension of the distance vector. We follow~\cite{park2022eigencontours,xu2019explicit} and uniformly quantize the 360-degree with an angle interval of $\Delta\theta=1^\circ$, hence $N=360$ throughout this paper.

The hierarchical encoding procedure involves subdividing the original object shape until a sufficiently regular region is obtained or the maximum depth, denoted by $D:=\lfloor\log_2 \overline{K}\rfloor$, is reached. At each level of subdivision, an angle-based contour descriptor is computed locally.  As depicted in~\Cref{alg:hie_encoding}, the current depth and solidity of $\mathcal{M}$ are examined first, and if the maximum depth has not been attained and the solidity is not greater than $\tau$ (where $\tau$ is set to 0.9\footnote{As the threshold for the desired solidity of subdivisions at termination, a higher $\tau$ indicates greater regularity or convexity. Generally, a solidity above 0.8 denotes a shape that is reasonably close to convex. For our experiments, we set $\tau=0.9$  to ensure a high degree of precision, aligning with standards for well-convex shapes.}), the shape is be subdivided. Since there is often no prior knowledge of how to create the partition, we propose to select one direction at a time to divide the shape into two refined components. This process is repeated until the termination conditions are satisfied. Specifically, viewing the object boundary points as two-dimensional data, we partition the data points according to the direction of the minimum data variance (denoted as $P$), or equivalently, the second principal direction of the data matrix. This direction passes through the mass center $\bm c_{\mathcal{M}}$ of $\mathcal{M}$. We write it as
\begin{equation}
     \mathcal{M}_1, \mathcal{M}_2 = \textsc{Split}(\mathcal{M},\bm c_{\mathcal{M}}, P),
\end{equation}
where $\mathcal{M}_1$ and $\mathcal{M}_2$ are the two resulting splits. This practice is similar to what is known as \emph{principal direction divisive partitioning (PDDP)}~\cite{boley1998principal}, with the aim of preserving the most spread-out information for subsequent use.

\begin{figure}[t]
    \centering
\includegraphics[width=.55\linewidth]{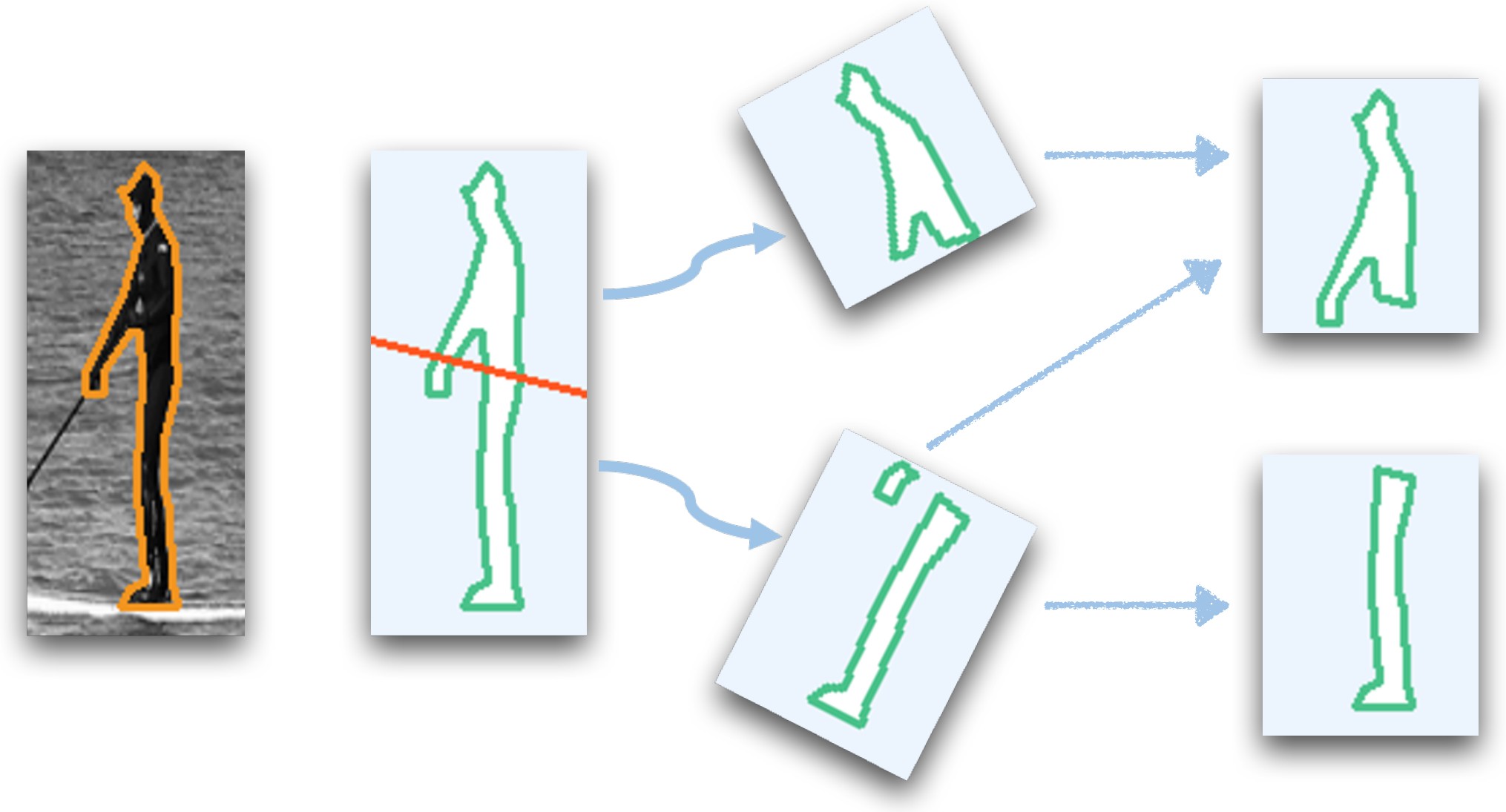}
\caption{\textbf{Reorganization of disconnected components.} Disconnected components can sometimes result from splitting the shape based on the direction of the least data variance. We handle the issue by retaining the component with the largest area and returning the other parts to the remainder of the partition.}
    \label{fig:reorg}
\end{figure}

One potential issue with the \textsc{Split} operation is that it can occasionally result in disconnected components in $\mathcal{M}_1$ and/or $\mathcal{M}_2$ (see~\Cref{fig:reorg}). In such cases, evaluating the solidity makes no sense, and finding a local contour that accurately describes the boundary becomes infeasible. To address this issue, we propose a \textsc{Reorg} operation. If $\mathcal{M}_1$ is disconnected, \textsc{Reorg} retains the component with the largest area in $\mathcal{M}_1$ and returns the remaining portions with smaller areas to $\mathcal{M}_2$. Note that $\textsc{Reorg}$ does not introduce new disconnected components in each part as it only involves the consolidation of the constituent parts, thereby preserving the connectivity of the original shape. After $\textsc{Reorg}$, $\mathcal{M}_1$ is guaranteed to be connected. If $\mathcal{M}_2$ also contains disconnected component, one can repeat the $\textsc{Reorg}$ operation on $\mathcal{M}_2$. Since $\textsc{Reorg}$ does not introduce new disconnected components, after the second $\textsc{Reorg}$, each final partition is ensured to contain exactly one connected component. We write the above process as 
\begin{equation}
    \mathcal{M}_1, \mathcal{M}_2 = \textsc{Reorg}(\mathcal{M}_1, \mathcal{M}_2).
\end{equation}
We can now formalize the validity of the procedure with the following proposition.
\begin{prop}
For a connected input $\mathcal{M}$, in each function call of \textsc{Hierarchical-Encoding}, \cref{line:m} will be executed at most twice. Then, the condition in~\cref{line:s} will not be satisfied, and therefore the while loop will be exited.
\end{prop}

Once the two refined subdivisions $\mathcal{M}_1$ and $\mathcal{M}_2$ are obtained with potentially higher regularity than $\mathcal{M}$, the previous subdivision procedure is repeated with an updated maximum hierarchical depth of $D-1$ for $\mathcal{M}_1$ and $\mathcal{M}_2$, respectively. The algorithm terminates whenever it detects a sufficiently regular shape or the predetermined depth is achieved, at which point we begin encoding a local contour to describe the shapes at the finest level. To locate its center $\bm c$, we adopt the mass center if it falls inside the shape; otherwise, we use the center of the largest inscribed circle. The distance vector $\bm r\in\mathbb{R}^{360}$ is uniformly sampled.

\Cref{alg:hie_encoding} can be regarded as a greedy approach for solving~\eqref{eq}. Though it is greedy, the subregions become increasingly regular as the subdivision progresses. This is because the shapes are split into smaller areas using simple lines, which serve as the new boundaries of those subareas, thus making it more amenable to representation by simple local contours. For instance, in~\Cref{fig:dog}, both objects are hierarchically encoded using~\Cref{alg:hie_encoding} with depths of $1,2$ and 5. As the object boundary of the lady is sufficiently regular, the algorithm terminates by using one single contour to describe the shape in all scenarios, which proves to be effective. For the challenging case of the man, as the maximum allowable depth increases, the algorithm continues to subdivide the region. Remarkably, even without prior geometric information, local contours emerge automatically around challenging areas, demonstrating the adaptive subdivision capability of PDDP.

\begin{figure}[t]
    \centering
\includegraphics[width=.45\linewidth]{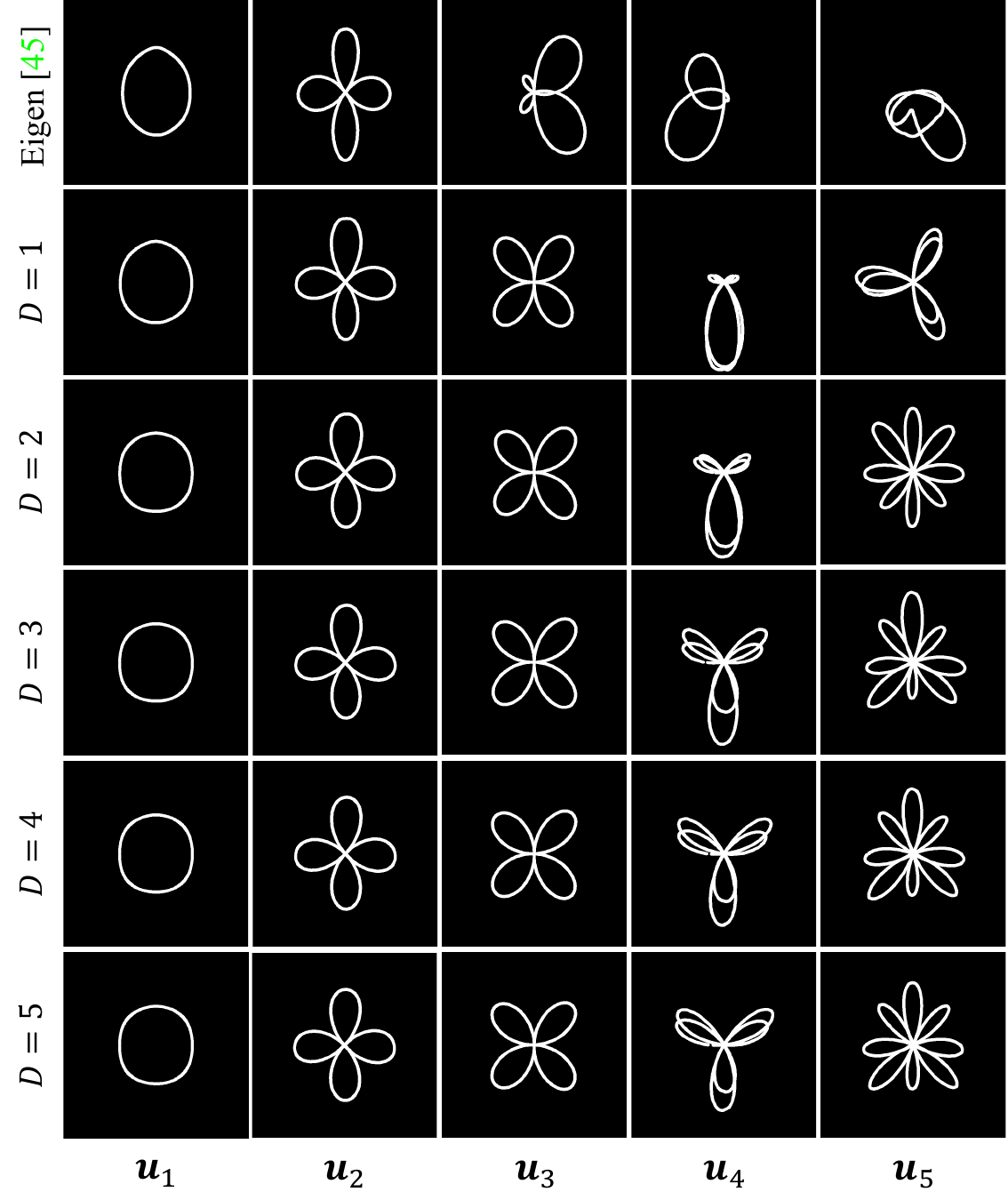}
    \caption{\textbf{Visualization of the five basis vectors in $\bm U^\star$ with $M=5$.} Top to bottom: we increase depth $D$ for encoding all instances in the COCO2017 dataset~\cite{lin2014microsoft}.}
    \label{fig:vis-eigen}
\end{figure}

\subsection{Robust subspace projection} \label{subsec:factorization}

The interdependence of object shapes motivates the utilization of low-rank subspace projection to effectively capture such structural patterns.  We denote by $L$ the number of objects in a training set, and we use $\widetilde L$ to denote the total number of local contours after encoding (clearly, $\widetilde L\ge L$). The contour matrix $\bm A =[\bm r_1,\bm r_2,\cdots, \bm r_{\widetilde L}]\in\mathbb{R}^{N\times \widetilde L}$ is constructed by concatenating the distance vectors of the local contours. \cite{park2022eigencontours} simply applies SVD on $\bm A$ to compute a $M$-dimensional subspace $\mathcal{S}$ with $M\ll N$. However, it is well-known that the classic $\ell_2$-based SVD solution is very sensitive to outliers \cite{lerman2018overview,lerman2018fast,tsakiris2015dual,ding2021dual,ding2021dualhyper}, a category that includes the distance vectors of irregular object shapes.

We aim at robustly estimating the underlying low-dimensional subspace in the presence of outliers, known as the \emph{Robust Subspace Recovery (RSR)} problem~\cite{lerman2018overview, ding2021subspace,ding2019noisy,zhu2019linearly}. In this regard, we utilize the Fast Median Subspace (FMS)~\cite{lerman2018fast} approach that efficiently finds a basis $\bm U^\star$ for the subspace $\mathcal{S}$ by solving the non-convex problem of least absolute loss:
\begin{align}\label{eq:fms}
    \min_{\bm U\in\mathbb{O}(N,M)}
\sum\nolimits_{j=1}^{\widetilde L}\|(\bm I -\bm U \bm U^\top)\bm r_j\|_2,
\end{align}
where $\mathbb{O}(N,M):=\{\bm U\in\mathbb{R}^{N\times M}: \bm U^\top \bm U  = \bm I_{M}\}$ is the set of orthonormal matrices. It minimizes the sum of absolute deviations  from all the data points to their projections onto $\mathcal{S}$. Due to the geometric meaning of minimizing the least absolute deviations, it essentially estimates a ``median" basis of the underlying $M$-dimensional subspace, as opposed to a ``mean" one, and is therefore more robust.

\subsection{Basis-sharing conversion}\label{subsec:conversion}

Once the the $M$-dimensional robust basis $\bm U^\star$ is computed by solving~\eqref{eq:fms},
we are able to efficiently approximate the distance vectors using subspace projection. As $\bm U^\star$ is learned from $\bm A$, which contains the local contours of all the objects, we use the universal basis $\bm U^\star$ for reconstructing all of the local contours spread over different objects.

For the $j$-th object, $j=1,\cdots, L$, we denote by $\bm R^{(j)}=[\bm r_1^{(j)},\bm r_2^{(j)},\cdots,\bm r_{K_j}^{(j)}]$ the matrix containing $K_j$ distance vectors computed by~\Cref{alg:hie_encoding} ($\sum_{j=1}^L K_j=\widetilde L$). We now have the low-dimensional projection of $\bm R^{(j)}$, calculated by
\begin{align}\label{eq:share}
   \widehat{\bm R}^{(j)} = \bm U^\star {\bm U^\star}^\top \bm R^{(j)} = \bm U^\star \bm \Omega^{(j)},
\end{align}
where $\bm \Omega^{(j)}={\bm U^\star}^\top \bm R^{(j)}\in\mathbb{R}^{M\times K_j}$ is the coefficient matrix. Each $N$-dimensional distance vector is approximated by an $M$-dimensional row vector of $\bm \Omega^{(j)}$ that lies in $\mathcal{S}$. In practice, since $\bm U^\star$ is fixed after solving~\eqref{eq:fms}, one only needs to store the coefficient matrix $\bm \Omega^{(j)}$ to approximate the distance vectors. When combined with the centers, we are able to efficiently and faithfully reconstruct the local contours of each object. Finally, we retrieve the object boundary by using the outer edge of the union of the object's local contours.

\begin{figure}[t]
    \centering
\includegraphics[width=.5\linewidth]{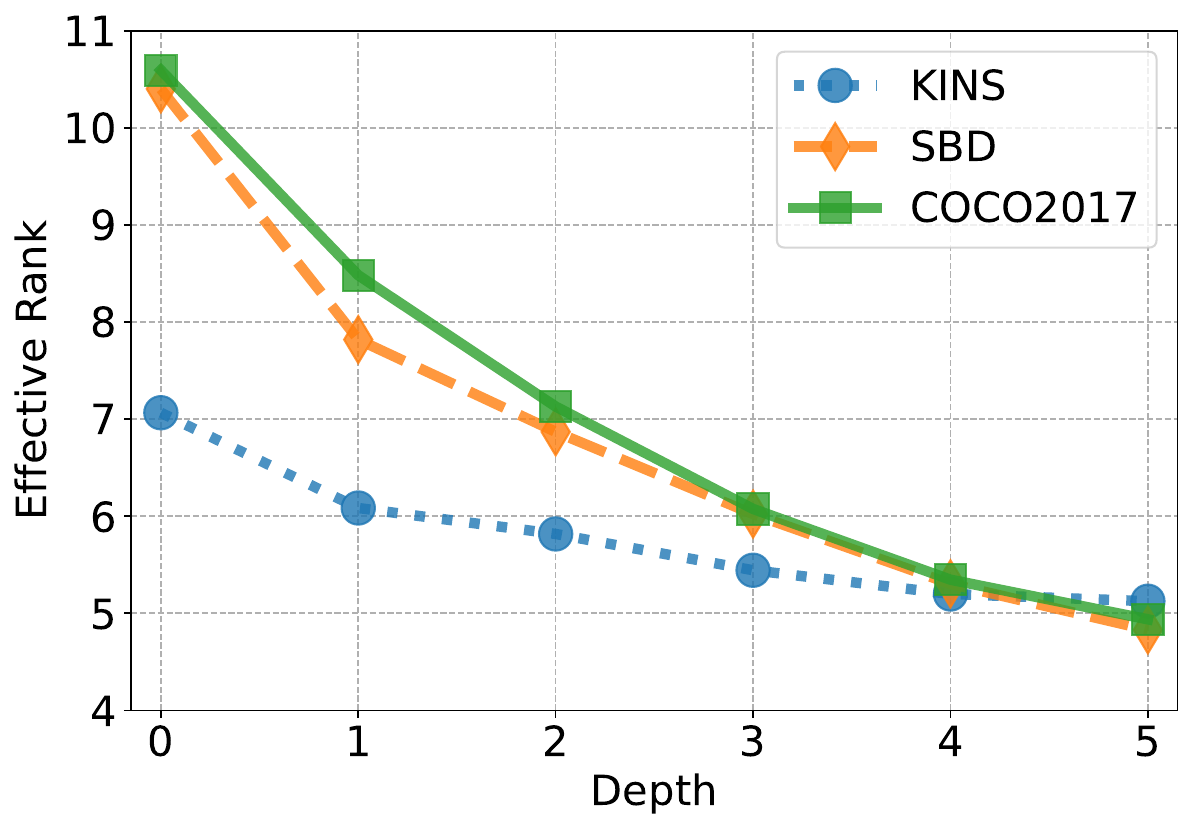}
\caption{\textbf{Effective rank of the contour matrix constructed with different hierarchical depths $D$.} As $D$ increases, the effective rank of the contour matrix decreases, suggesting more correlation among the contours.}
    \label{fig:rank}
\end{figure}

\subsection{Analysis and examples}\label{subsec:example}

\paragraph{Structural patterns from hierarchical encoding.}
We first examine the basis $\bm U^\star$ learned from the contour matrix, focusing on how the different values of hierarchical encoding depth $D$ affect the behavior of $\bm U^\star$. Towards that end, we use COCO2017 dataset~\cite{lin2014microsoft}, construct the contour matrix with the universal set of all instances encoded, and recover a five-dimensional subspace by solving~\eqref{eq:fms} with $M=5$. In~\Cref{fig:vis-eigen}, we visualize the five basis vectors (also eigenvectors) of $\bm U^\star$ associated with different values of $D$, and denote them as $\{\bm u_i\}_{i=1}^5$ in a descending order. 

Recall that Eigen~\cite{park2022eigencontours} uses a single global contour representation, thereby corresponding to the case where $D=0$. As $D$ grows, more local contours are developed to represent the subdivisions. Notably, one can observe in~\Cref{fig:vis-eigen} that, from top to bottom, the $\bm u_i$'s exhibit increasingly more symmetrical patterns. This is because, as the hierarchical encoding level rises,\emph{~\Cref{alg:hie_encoding} adaptively splits the shape into more regularly shaped subdivisions for encoding, resulting in a more regularized representation in a low-dimensional subspace}. Also, the most regular representations in the last row offer us with better interpretations for the basis vectors: $\bm u_1$ regulates the overall size of an object, resembling a circle with equal distance in all directions; $\bm u_2$ governs the additional horizontal and vertical size, while $\bm u_3$ controls the additional diagonal size; $\bm u_4$ and $\bm u_5$ provide more detailed information uniformly across other directions. \emph{Note that such structural patterns cannot be easily inferred from the first row, where a single contour is used, showing the benefit of the multiple local hierarchical representation.}

\begin{figure}[t]
    \centering
\includegraphics[width=.95\linewidth]{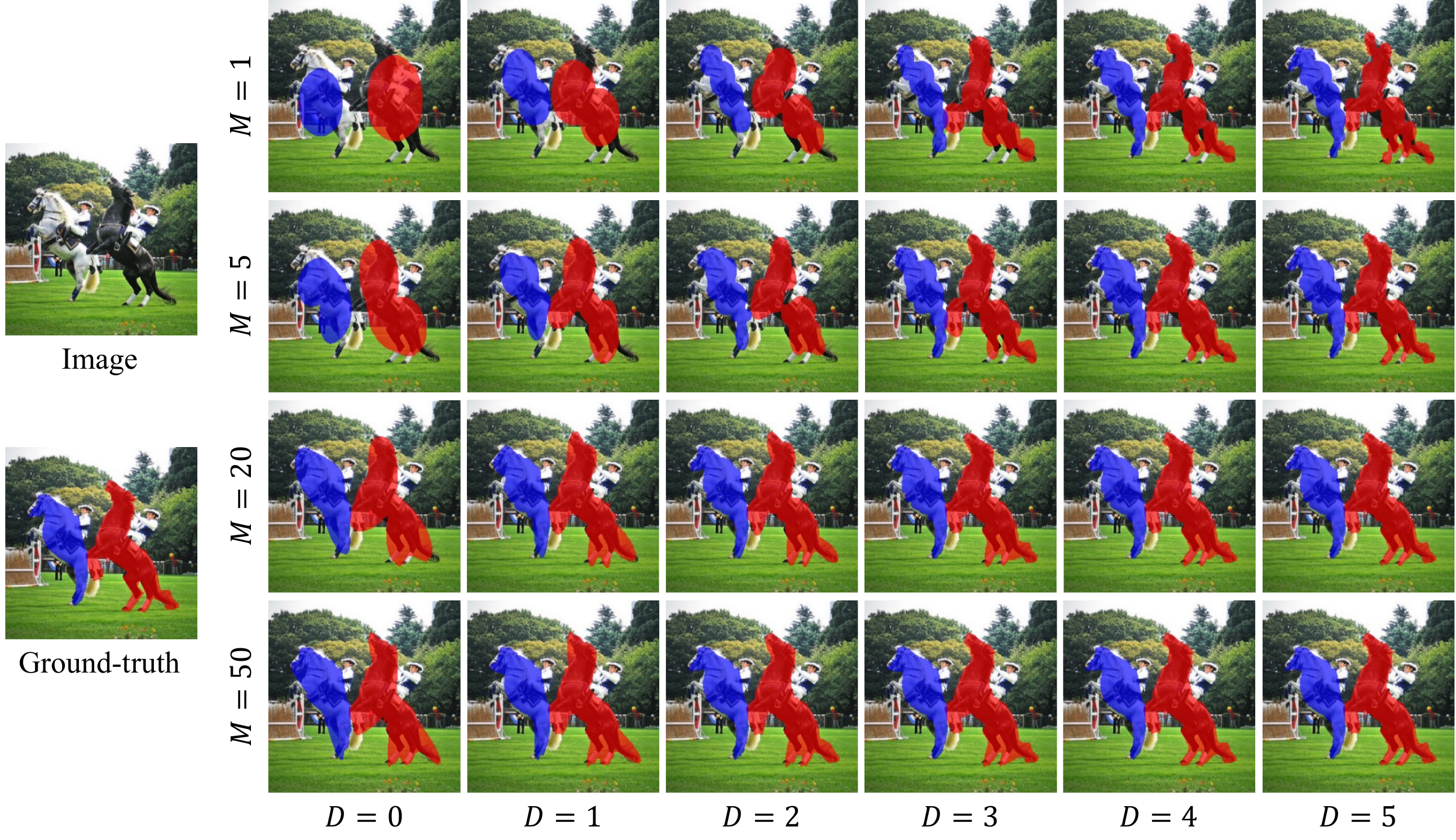}
    \caption{\textbf{Visualization of AdaContour reconstructions with different values of encoding depth $D$ and projection dimension $M$.}}
    \label{fig:horse}
\end{figure}

\paragraph{Effective rank of the contour matrix.}
We validate further the regularized representation by calculating the effective rank of the contour matrix as the numerical sparsity~\cite{lopes2013estimating} of the singular values of $\bm A$ (denoted by $\bm s(\bm A)$), which is defined as $\|\bm s(\bm A)\|_1^2/\|\bm s(\bm A)\|_2^2$. \Cref{fig:rank} plots the effective rank of $\bm A$ constructed with different values of hierarchical depth $D$ for encoding all instances in KINS~\cite{qi2019amodal}, SBD~\cite{hariharan2011semantic}, and COCO2017~\cite{lin2014microsoft} datasets. It shows that the effective rank of $\bm A$ continues to decrease as $D$ grows, indicating that the local contours encoded with a higher hierarchical level lie in a lower dimensional subspace, which, from another perspective, verifies that the shapes of the local subdivisions become more structural due to their increased correlation. We also see that the rank associated with the KINS dataset is often lower than that of SBD and COCO2017 owing to the fact that all the objects in KINS are vehicles and people, which exhibit strong correlations by nature.

\paragraph{Expressive capability analysis.}
We conduct an expressive capability analysis by testing on an example from the COCO2017 dataset, with varying settings of hierarchical encoding depth ($D$) and projection dimension ($M$). \Cref{fig:horse} displays the results where $D$ increases from left to right, and $M$ increases from top to bottom. One can see that when a single global contour is used (the first column), even with a rank-20 approximation, the object boundary cannot be accurately characterized, especially for the right horse. In particular, for the upper-left figure with $M=1$ and $D=0$, each horse is represented by a single ellipse centered on the midsection. As $D$ increases from 0 to 1, two local contours representing the upper and lower portions of the body develop on each horse, and as $D$ further increases, more local contours are produced at critical areas to appropriately represent the shape. \emph{Intriguingly, even with only one basis vector ($M=1$), the general skeletons of the horses are depicted in the upper-right figure with $D=5$.} Moreover, as $M$ grows, more accurate contour representations are constructed for all instances. We remark that, a large encoding depth $D$ can compensate for a relatively small $M$ used in the low-rank approximation, e.g., the case of $D=5,M=5$ corresponds to a more precise representation than the case of $D=0, M=20$. 
Additionally,  the basis-sharing conversion~\eqref{eq:share} ensures that the same set of basis, i.e., $\bm U^\star$, is used to efficiently reconstruct all the local contours with a single matrix multiplication.

\begin{table}[]
\caption{\textbf{IOUs between the ground-truth and the result of different contour descriptors.} The associated datasets are KINS, SBD, and COCO2017 (from top to bottom).}
\label{tab:iou}
\footnotesize
\centering
\begin{tabular}{c|cc|ccccc}
\toprule
 & \multirow{2}{*}{\begin{tabular}[c]{@{}c@{}}ESE\\ \cite{xu2019explicit}\end{tabular}} & \multirow{2}{*}{\begin{tabular}[c]{@{}c@{}}Eigen\\ \cite{park2022eigencontours}\end{tabular}} & \multicolumn{5}{c}{Ours w/ different choices of $D$ {\scriptsize $\color{red}\uparrow\! \text{over SVD}$}}                         \\
$M$   &                     &                        & 1     {\scriptsize $\color{red}\uparrow \! 0.5$}                 & 2            {\scriptsize $\color{red}\uparrow \! 0.3$}         & 3          {\scriptsize $\color{red}\uparrow \! 0.3$}           & 4      {\scriptsize $\color{red}\uparrow \! 0.0$}               & 5      {\scriptsize $\color{red}\uparrow \! 0.0$}               \\
\midrule
1                       &    42.0                 &           56.0              &            63.2         &         65.3              &          \underline{66.5}             &        \textbf{66.8}             &            \textbf{66.8}          \\
5                       &        53.0              &          72.6              &           78.3           &         79.2             &        80.3              &        \underline{80.4}              &          \textbf{80.5}            \\
20                      &         74.0             &          81.1              &        86.0              &           86.8           &       \underline{87.3}               &          \textbf{87.4}            &              \textbf{87.4}        \\
50                      &       80.0               &           82.5             &          87.3            &         87.9             &       88.2               &           \underline{88.3}           &             \textbf{88.4}         \\
100                     &         82.3             &         82.5               &      87.1                &        87.6              &       \underline{88.0}               &            \textbf{88.1}          &     \textbf{88.1}                 \\
360                     &            87.1          &              87.2          &         89.5             &            90.2          &     90.5                 &         \underline{90.6}             &              \textbf{90.7}        \\
\bottomrule
\end{tabular}
\\

\begin{tabular}{c|cc|ccccc}
\toprule
 & \multirow{2}{*}{\begin{tabular}[c]{@{}c@{}}ESE\\ \cite{xu2019explicit}\end{tabular}} & \multirow{2}{*}{\begin{tabular}[c]{@{}c@{}}Eigen\\ \cite{park2022eigencontours}\end{tabular}} & \multicolumn{5}{c}{Ours w/ different choices of $D$ {\scriptsize $\color{red}\uparrow\! \text{over SVD}$}}                         \\
$M$   &                     &                        & 1     {\scriptsize $\color{red}\uparrow \! 1.0$}                 & 2            {\scriptsize $\color{red}\uparrow \! 0.7$}         & 3          {\scriptsize $\color{red}\uparrow \! 0.1$}           & 4      {\scriptsize $\color{red}\uparrow \! 0.0$}               & 5      {\scriptsize $\color{red}\uparrow \! 0.0$}               \\
\midrule
1                       &    51.2                  &           50.1              &            63.1           &         67.4              &          69.9             &        \underline{71.3}               &            \textbf{72.0}           \\
5                       &       63.3               &        69.4                &       76.3               &              80.5        &       83.3               &         \underline{84.8}             &   \textbf{85.7}                   \\
20                      &         82.2             &          84.5              &       89.1               &             91.2         &      92.5                &           \underline{93.2}           &  \textbf{93.6}                    \\
50                      &            87.3          &           88.0             &        91.9              &         93.3             &      94.2                &          \underline{94.7}            &    \textbf{94.9}                  \\
100                     &          88.5            &          88.7              &       92.3               &       93.5               &      94.3                &          \underline{94.8}            &   \textbf{95.0}                   \\
360                     &             89.6         &        90.0                &         92.7             &            93.9          &       94.7               &           \underline{95.2}           &  \textbf{95.4}                    \\
\bottomrule
\end{tabular}
\\

\begin{tabular}{c|cc|ccccc}
\toprule
 & \multirow{2}{*}{\begin{tabular}[c]{@{}c@{}}ESE\\ \cite{xu2019explicit}\end{tabular}} & \multirow{2}{*}{\begin{tabular}[c]{@{}c@{}}Eigen\\ \cite{park2022eigencontours}\end{tabular}} & \multicolumn{5}{c}{Ours w/ different choices of $D$ {\scriptsize $\color{red}\uparrow\! \text{over SVD}$}}                         \\
$M$   &                     &                        & 1     {\scriptsize $\color{red}\uparrow \! 1.7$}                 & 2            {\scriptsize $\color{red}\uparrow \! 0.5$}         & 3          {\scriptsize $\color{red}\uparrow \! 0.2$}           & 4      {\scriptsize $\color{red}\uparrow \! 0.0$}               & 5      {\scriptsize $\color{red}\uparrow \! 0.0$}               \\
\midrule
1                       &    48.8                  &           48.4              &            58.3           &         61.8              &          63.6             &        \underline{64.4}               &            \textbf{64.8}           \\
5                       &        61.0              &        67.0                &          73.0            &          75.5            &     78.3                 &       \underline{79.2}               &               \textbf{79.6}       \\
20                      &             79.0         &          80.8              &       84.0               &         85.5             &      86.4                &           \underline{86.7}           &     \textbf{86.8}                 \\
50                      &         83.1             &        83.6                &          86.3            &       87.3               &    87.9                  &         \underline{88.1}             &          \textbf{88.2}            \\
100                     &     83.8                 &          83.9              &       86.5               &           87.4           &             87.9         &      \underline{88.1}                &         \textbf{88.2}             \\
360                     &          88.5            &          88.7              &     91.2                 &          92.1            &      \underline{92.6}                &       \textbf{92.9}               &        \textbf{92.9}              \\
\bottomrule
\end{tabular}
\end{table}

\section{Experiments}

\paragraph{Datasets.} We use three datasets: (i) KINS~\cite{qi2019amodal} that augments KITTI~\cite{geiger2012we} for amodal instance segmentation and contains 7,474/7,517 images with 7 categories for training/testing; (ii) SBD~\cite{hariharan2011semantic} that contains 5,623/5,732 images from the PASCAL VOC challenge~\cite{everingham2010pascal} with 20 categories for training/testing; and (iii) COCO2017~\cite{lin2014microsoft} that contains 118K/5K images with 80 categories for training/testing.

\paragraph{Comparative assessment.} We compare AdaContour to the recently proposed Eigencontours~\cite{park2022eigencontours} (abbreviated as ``Eigen") and ESE-Seg~\cite{xu2019explicit} (abbreviated as ``ESE"), both of which employ a single global contour representation. Note that conventional contour descriptors such as Polynomial~\cite{cinque1998shape} and Fourier~\cite{zhang2001comparative} have been shown to be significantly less effective than Eigen in~\cite{park2022eigencontours}. Therefore, they are excluded in this comparison due to space constraints. \Cref{tab:iou} compares the IOUs between ground-truth object masks and the results of different approaches. In ESE, $M$ is the number of Chebyshev polynomial coefficients for fitting the distance vector in a contour, whereas in Eigen, $M$ is the rank of the eigencontour space calculated by SVD. We use all instances in all categories of the datasets. 

\begin{figure}[t]
    \centering
\includegraphics[width=.98\linewidth]{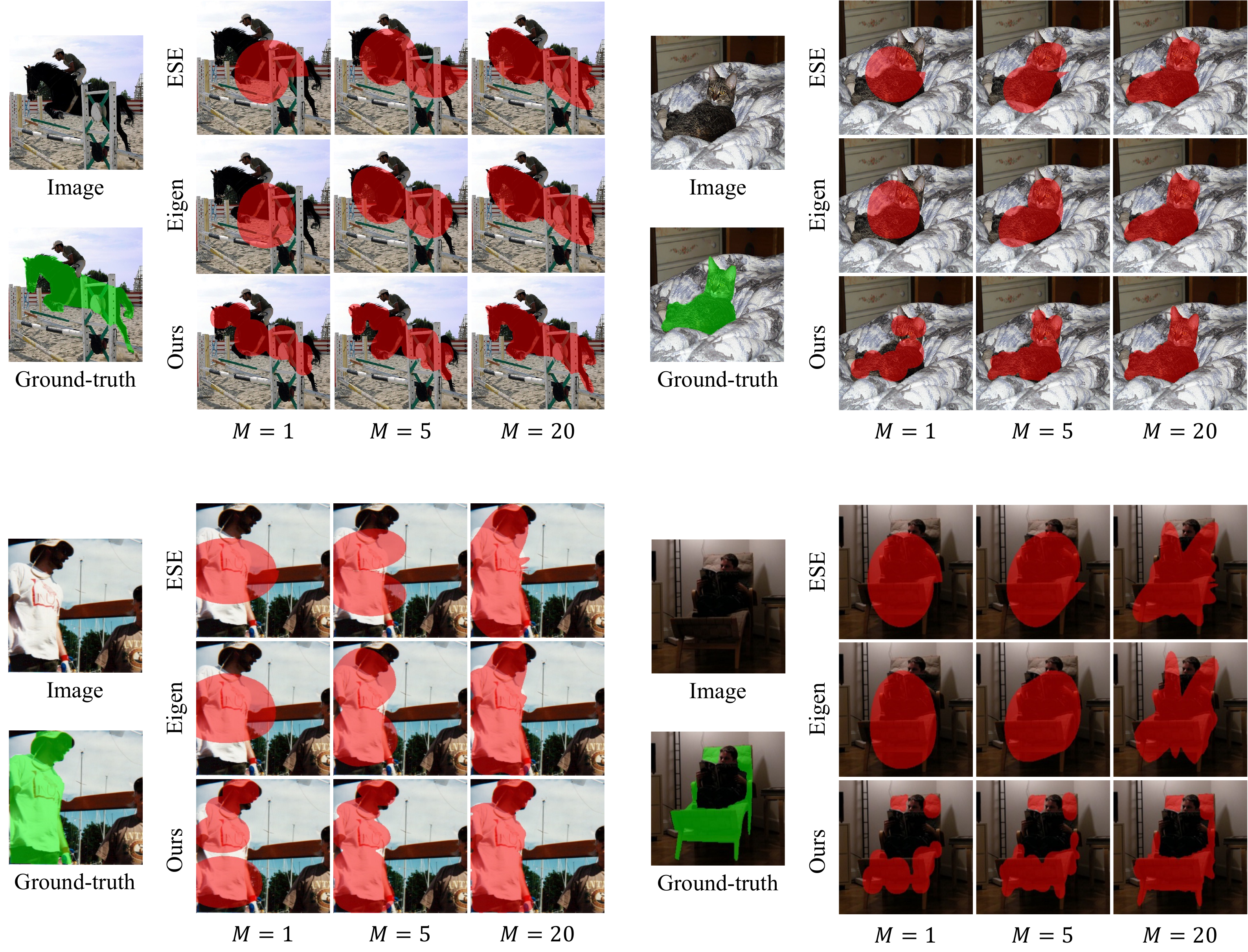}
    \caption{\textbf{Qualitative comparison ($D=3$ for ours).}}
    \label{fig:sofa}
\end{figure}

We observe that AdaContour with multiple local representations ($D>0$) \emph{performs the best on all three datasets at every $M$}. Specifically, with $D=1$ (maximum of two local contours), AdaContour achieves IOUs of 86.0, 89.1, and 84.0 on KINS, SBD and COCO2017 at $M=20$, surpassing ESE and Eigen with $M=100$. With increasing $D$, our method's advantage becomes more evident in terms of accuracy, albeit with the cost of a less compact representation. \emph{We highlight that, on KINS and SBD, our method with $D=1,M=50$ beats ESE and Eigen with $M=360$}, i.e., the original encoding without low-rank approximation, demonstrating its superiority. The compensatory effect of a high  $D$ value for a low $M$ value is  also evident from~\Cref{tab:iou}: the setting of $D=5,M=20$ consistently outperforms that of $D=1,M=100$. Finally, we mark in red the overall improvements achieved by using RSR to solve~\eqref{eq:fms} as compared to~\cite{park2022eigencontours} that uses pure SVD. It shows that RSR generates more accurate results, especially at $D=1$. As $D$ grows, the advantage of using RSR reduces since a more regularized representation contributes to a more structural subspace, and a pure SVD can already yield decent results.

\Cref{fig:sofa} compares qualitatively the contour representations produced by different descriptors, where we use $D=3$ in our method. Neither ESE nor Eigen can faithfully represent the object in the provided examples with a single contour. In contrast, AdaContour reliably represents the object in all $M$ configurations. Notably, the ground-truth object region in the sofa case is highly irregular, but our method performs significantly better than the two competitors, even at $M=1$, due to its ability to capture intricate local curvature information.

\begin{table*}[t]
\caption{\textbf{Quantitative comparisons on SBD validation data.}}\label{tab:ins}
\footnotesize
\centering
\begin{tabular}{l|ccc}
\toprule
  & $\text{AP}_{50}$ & $\text{AP}_{75}$& $\text{AP}_{\mathcal{F}}$  \\
\midrule                
PolarMask~\cite{xie2020polarmask}    &  \hspace{.12in}50.11\hspace{.12in}    &  \hspace{.12in}14.50\hspace{.12in} &  \hspace{.12in}25.78\hspace{.12in}      \\
ESE~\cite{xu2019explicit}            &52.14       &20.48     &   27.37  \\
Eigen~\cite{park2022eigencontours}           &56.47     &29.35     &  \textbf{35.30}    \\
\midrule
Proposed ($M=10$)\hspace{.12in} &56.69   &28.44     &  32.16  \\
Proposed ($M=20$) & \underline{57.57}      & \underline{31.88}     & \underline{34.16}    \\
Proposed ($M=30$) & \textbf{58.46}      & \textbf{32.64}     &  33.51   \\
\bottomrule
\end{tabular}
\end{table*}

\begin{figure}[t]
    \centering
\includegraphics[width=.6\linewidth]{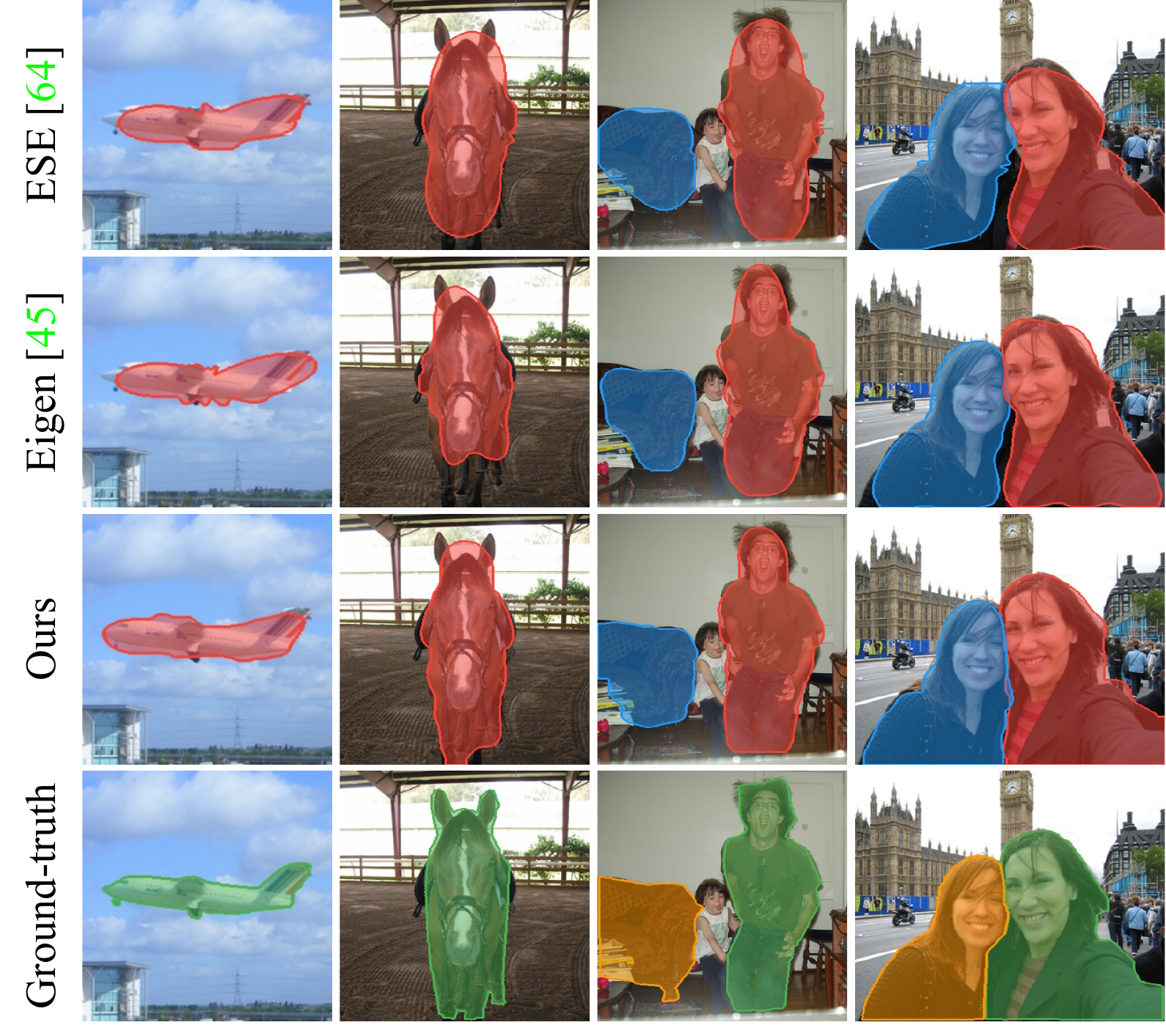}
    \caption{\textbf{Qualitative comparisons on SBD validation data.}}
    \label{fig:demo}
\end{figure}

\paragraph{Instance segmentation.} The angle-based contour descriptors can be explicitly incorporated into object detection frameworks and enable one-stage instance segmentation~\cite{xie2020polarmask,xu2019explicit,park2022eigencontours}. To validate AdaContour's suitability for instance segmentation, we encode each object with exactly two local contours using $D=1$ and $\tau=0.99$. Then, we utilize the YOLOv3~\cite{redmon2018yolov3} detector, and follow the practice in~\cite{xu2019explicit}. In addition to using the original YOLOv3 output for bounding box regression and object classification, we add extra components to regress a $2M$-dimensional coefficient vector and a 4-dimensional position vector of two centers for each object, both of which are encoded by AdaContour. Finally, the shape mask of an object is reconstructed by simultaneously calculating the two local contours using~\eqref{eq:share}. \Cref{tab:ins} compares the instance segmentation results on the SBD validation dataset. We follow~\cite{park2022eigencontours} and report the average precision (AP) scores for two IOU thresholds of 0.5 and 0.75, as well as a $\mathcal{F}$ score threshold of 0.3. For $\text{AP}_{50}$ and $\text{AP}_{75}$, our method with $M\in\{20,30\}$ is better than the other competitors. Note that our method with $M=10$ already outperforms Eigen with $M=20$ when $D=1$ is used. \Cref{fig:demo} gives visual comparisons showing that the regression with the encoding from two local contours delivers more accurate results.

\begin{figure}[t]
\centering
\begin{minipage}{0.45\linewidth}
    \centering
    \includegraphics[width=\linewidth]{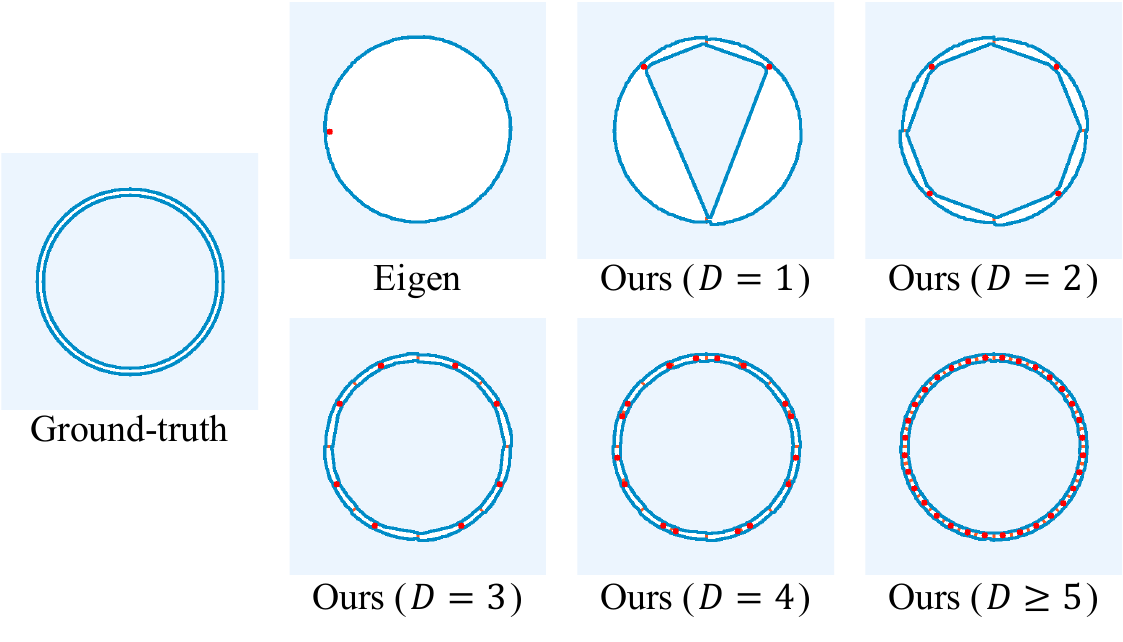}
    \caption{\textbf{A difficult example of a {ring} shape}.}
    \label{fig:donut}
\end{minipage} \ \ 
\begin{minipage}{0.50\linewidth}
    \centering
    \footnotesize
    \captionof{table}{\textbf{Performance and resource consumption with different choices of $D$.}}
    \label{tab:performance}
    \begin{tabular}{c|c|ccccc}
    \toprule
     &  \multirow{2}{*}{\begin{tabular}[c]{@{}c@{}}Eigen\end{tabular}} & \multicolumn{5}{c}{Ours w/ different choices of $D$} \\
       &                                           & 1                   & 2               & 3                 & 4            & $\ge$5                  \\
    \midrule
    IOU $\uparrow$                                    &          0.12            &           0.19       &         0.72              &          0.85             &        0.89             &            \textbf{0.95}          \\
    CPU-time (s) $\downarrow$                                 &          0.25             &           \textbf{0.20}           &         0.29             &    0.51              &       0.69              &          1.02            \\
    Memory (MB) $\downarrow$                                 &          \textbf{3.6}            &        8.9          &    12.7            &    15.3            &   25.5            &         35.5         \\
    \bottomrule
    \end{tabular}
\end{minipage}
\end{figure}

\paragraph{Discussion.}
AdaContour operates by computing only the \emph{necessary} subdivisions and \emph{stopping early} when it identifies a regular shape. As a result, for regular shapes, it performs {comparably}, or even \emph{faster} than conventional algorithms due to our optimized implementation. For extremely irregular shapes, such as a ring (see~\Cref{fig:donut}), increasing $D$ significantly improves the accuracy. While this leads to a slight increase in computation time, it {remains highly efficient} (see~\Cref{tab:performance}): for \emph{all $D\ge 5$, termination occurs $\sim$1s} (because of the early stopping), yielding an IOU of 0.95, significantly higher than Eigen's result of 0.12. Furthermore, its memory footprint is at most $\sim$35.5~MB for this hard example, which is {negligible} for modern machines. These observations clearly suggest that AdaContour excels with a unique blend of efficiency and accuracy. Note that larger $D$ values only \textit{marginally} increase memory usage, which modern systems can {easily} accommodate. At $D= 5$, memory consumption $\le {35.5}$~{MB}, which is a \textit{fixed} upper limit for encoding shapes of {any} complexity from {any} dataset. This is because with $D= 5$, we can have up to 32 centers in $\mathbb{R}^2$ and radius vectors  in $\mathbb{R}^{360}$ using FP32 format. Thus, the memory requirements are consistent and predictable. 

We remark that our approach is specifically designed for irregular shapes to capture complex geometries, thus the benefits may not be as apparent for regular shapes. Consequently, the choice of $D$ primarily affects the accuracy when characterizing complex shapes, while the computation time for regular shapes remains nearly the same regardless of $D$ being large or small. Our experiments with the COCO2017 dataset (approximately 883K objects) show that an average of 4 local contours (corresponding to 
$D=2$) is sufficient, albeit that $D$ was set to 5 in experiments and can lead to a maximum of 32 local contours. In practice, we find that selecting $D=2$ typically yields highly satisfactory results compared to methods like Eigen, while still maintaining high efficiency.

Although a more systematic method for dividing shapes involves sampling from a skeleton, such methods, while providing a `thin' representation, often lose intricate contour details. In contrast, our method preserves such details. The simplicity of our current principal direction divisive partitioning, which directly operates on contour information, offers straightforward implementation and remains robust to boundary variations. Conversely, skeleton-based methods may introduce complexities due to additional sampling or processing needs. Compared to iterative methods like level sets and active contours, our descriptor-based approach offers efficiency and robustness to shape variations and noise, unlike snake models. Furthermore, AdaContour simplifies implementation and usage compared to geodesic active contour models.

\paragraph{Future directions.} Although the methods listed in~\Cref{tab:ins} are effective for instance segmentation when directly integrated into a detector, we recognize that such explicit approaches~\cite{xie2020polarmask,xu2019explicit,park2022eigencontours} are less competitive compared to implicit contour-based instance segmentation methods~\cite{liu2021dance,peng2020deep} that refine object boundaries \emph{end-to-end}. Encoding results like coefficients and centers can be challenging, especially with multiple contours. However, we propose a future development where an end-to-end pipeline uses the hierarchical representation in AdaContour as a form of strong regularization. This would construct the shape mask directly from multiple local contour predictions and apply final supervision on the mask rather than on intermediate encoding results. This approach could also address the vertex interlacing issues~\cite{liu2021dance} seen in vertex-based methods.

As a robust shape descriptor, AdaContour has potential applications beyond segmentation. Its hierarchical structure lends itself to shape deformation tasks~\cite{weng20062d,wang2015linear} by allowing manipulation of individual local contours while preserving the overall shape structure. The method's adaptability with different $D$ settings also makes it suitable for the registration and matching of shapes~\cite{van2011survey,belongie2002shape,veltkamp2001state}. Another potential direction involves leveraging generative models such as diffusion models~\cite{zhou2023dream,sohl2015deep,ho2020denoising} to learn the distribution of encoded local contours, opening avenues for graphic generation, particularly for complex shapes and geometries. Furthermore, it could aid in view synthesis tasks like neural rendering~\cite{bao2023insertnerf,zhu2023caesarnerf,bao2023and,mildenhall2021nerf}, providing strong regularization on specific objects during reconstruction. We believe this study provides a fresh perspective on developing more effective contour descriptors and broadens the scope for various downstream applications.

\section{Conclusions}

We presented an adaptive contour descriptor (AdaContour) based on hierarchical representations.  We exploited multiple local contours to desirably capture complicated geometric information, as opposed to the single global contour used by conventional approaches. A novel hierarchical encoding algorithm is designed for adaptively encoding the shape. Experiments revealed that, when combined with robust subspace learning and basis-sharing conversion, AdaContour is able to represent shapes more reliably.  Its performance has been further validated by the incorporation into instance segmentation framework.


\bibliographystyle{ACM-Reference-Format}
\bibliography{main}










\end{document}